\documentclass{article}

\usepackage[verbose=true,letterpaper]{geometry}
\usepackage[utf8]{inputenc}
\usepackage[T1]{fontenc}
\usepackage{url}
\usepackage{booktabs}
\usepackage{amsfonts}
\usepackage{nicefrac}
\usepackage{microtype}
\usepackage{lipsum}
\usepackage{graphicx}
\usepackage[numbers]{natbib}
\usepackage{doi}
\usepackage{textcomp}
\usepackage{amsmath,amssymb}
\usepackage[group-separator={ \, },separate-uncertainty=true]{siunitx}
\usepackage{color}
\usepackage{xcolor}
\usepackage{url}
\usepackage{fancyhdr}

\usepackage{listings}
\definecolor{codegreen}{rgb}{0,0.6,0}
\definecolor{codegray}{rgb}{0.5,0.5,0.5}
\definecolor{codepurple}{rgb}{0.58,0,0.82}
\definecolor{backcolour}{rgb}{0.95,0.95,0.95}
\lstdefinestyle{mystyle}{
    backgroundcolor=\color{backcolour},
    commentstyle=\color{codegreen},
    keywordstyle=\color{magenta},
    numberstyle=\tiny\color{codegray},
    stringstyle=\color{codepurple},
    basicstyle=\ttfamily\footnotesize,
    breakatwhitespace=false,
    breaklines=true,
    captionpos=b,
    keepspaces=true,
    numbers=left,
    numbersep=5pt,
    showspaces=false,
    showstringspaces=false,
    showtabs=false,
    tabsize=2
}
\usepackage{hyperref}
\usepackage[capitalise]{cleveref}

\newcommand{\etal}{\textit{et al}. }
\newcommand{\ie}{\textit{i}.\textit{e}., }
\newcommand{\eg}{\textit{e}.\textit{g}., }
\newcommand{\email}[1]{\href{mailto:#1}{#1}}
\newcommand{\supsep}{\textsuperscript{,}}
\newcommand{\inst}[1]{\footnotemark[#1]~}
\newcommand{\orcidID}[1]{#1}

\newcommand{\ncars}{\texttt{N-CARS}}
\newcommand{\nmnist}{\texttt{N-MNIST}}
\newcommand{\shapes}{\texttt{\detokenize{shapes_translation}}}
\newcommand{\shapesEasy}{\texttt{\detokenize{shapes_translation-90}}}
\newcommand{\shapesHard}{\texttt{\detokenize{shapes_translation-30}}}
\newcommand{\shapesEasyHard}{\texttt{\detokenize{shapes_translation-30/90}}}
\newcommand{\mnist}{\texttt{MNIST}}

\newcommand{\DenseSep}{DenseSep}

\newcommand{\accum}{\mathrm{Acc}}
\newcommand{\tout}{T_{\mathrm{out},i}}
\newcommand{\synweight}{w}

\newcommand{\state}{V}
\newcommand{\current}{I}
\newcommand{\spikes}{S}
\newcommand{\thresh}{V_\mathrm{th}}
\newcommand{\sumsmall}{\textstyle\sum}
\newcommand{\memleak}{\beta}
\newcommand{\synleak}{\alpha}
\newcommand{\taumem}{\tau_\mathrm{mem}}
\newcommand{\tausyn}{\tau_\mathrm{syn}}
\newcommand{\timestep}{t_d}
\newcommand{\ntimesteps}{n_\mathrm{t}}
\newcommand{\gradscale}{\lambda}
\newcommand{\outint}{\Delta T}
\newcommand{\nouts}{n_\mathrm{out}}
\newcommand{\nstacks}{n_\mathrm{s}}
\newcommand{\nblockstack}{n_\mathrm{l}}
\newcommand{\relu}{\mathrm{ReLU}}
\newcommand{\conv}{\mathrm{C}}
\newcommand{\dropout}{\mathrm{Drop}}
\newcommand{\batchnorm}{\mathrm{BN}}
\newcommand{\pool}{\mathrm{P}}
\newcommand{\targetrate}{r_\mathrm{target}}
\newcommand{\lossspikes}{L_\mathrm{spikes}}
\newcommand{\lr}{\eta}

\newcommand{\numeps}{n_\mathrm{e}}
\newcommand{\lambdaspikes}{\lambda_{s}}
\newcommand{\lambdaout}{\lambda_{\mathrm{out}}}
\newcommand{\nlayersperblock}{N_\mathrm{l}}
\newcommand{\nblocks}{N_\mathrm{b}}
\newcommand{\growthfactor}{g}
\newcommand{\minsize}{s_\mathrm{min,i}}
\newcommand{\maxsize}{s_\mathrm{max,i}}
\newcommand{\aspratio}{r_\mathrm{i}}
\newcommand{\lossprior}{L_\mathrm{priors}}
\newcommand{\meaniou}{\mu_\mathrm{IoU}}
\newcommand{\detectionthresh}{t_\mathrm{d}}
\newcommand{\npriorsthresh}{N_\mathrm{priors>\detectionthresh}}
\newcommand{\npriors}{N_\mathrm{priors}}
\newcommand{\nchannels}{c_\mathrm{f}}
\newcommand{\fmapheight}{h_\mathrm{f}}
\newcommand{\fmapwidth}{w_\mathrm{f}}

\newcommand{\titlefull}{Hybrid SNN-ANN: Energy-Efficient Classification and Object Detection for Event-Based Vision}
\title{\titlefull}
\newcommand{\titlerunning}{Hybrid SNN-ANN: Energy-Efficient Event-Based Vision}
\hypersetup{
pdftitle={Hybrid SNN-ANN: Energy-Efficient Classification and Object Detection for Event-Based Vision},
pdfsubject={cs.CV},
pdfauthor={Alexander Kugele},
pdfkeywords={Computer Science, Event-based Cameras, Spiking Neural Networks, Artificial Neural Networks},
}
\newcommand{\headeright}{A. Kugele \etal}
\fancyheadoffset{0pt}
\makeatletter
\renewcommand{\@maketitle}{%
  \vbox{%
    \hsize\textwidth
    \linewidth\hsize
    \vskip 0.1in
    \centering
    {\LARGE\sc \@title\par}
    \vskip 0.1in
    \def\And{%
      \end{tabular}\hfil\linebreak[0]\hfil%
      \begin{tabular}[t]{c}\bf\rule{\z@}{24\p@}\ignorespaces%
    }
    \def\AND{%
      \end{tabular}\hfil\linebreak[4]\hfil%
      \begin{tabular}[t]{c}\bf\rule{\z@}{24\p@}\ignorespaces%
    }
    \begin{tabular}[t]{c}\bf\rule{\z@}{24\p@}\@author\end{tabular}%
  \vskip 0.4in \@minus 0.1in \center{\@date}   \vskip 0.2in
  }
}
\makeatother

\date{}

\author{

\textbf{Alexander~Kugele}\textsuperscript{*}\inst{1}\supsep{}\inst{2} \\ \email{alexander.kugele@de.bosch.com} \\ \orcidID{0000-0002-7482-8973}
\and

\textbf{Thomas~Pfeil}\inst{1} \\ \email{thomas.pfeil@de.bosch.com}
\and

\textbf{Michael~Pfeiffer}\inst{1} \\ \email{michael.pfeiffer3@de.bosch.com} \\ \orcidID{0000-0001-7159-3622}
\and

\textbf{Elisabetta~Chicca}\inst{2}\supsep{}\inst{3} \\ \email{e.chicca@rug.nl} \\ \orcidID{0000-0002-5518-8990}
}

\footnotetext[1]{Bosch Center for Artificial Intelligence, 71272 Renningen, Germany}
\footnotetext[2]{Bio-Inspired Circuits and Systems (BICS) Lab, Zernike Inst Adv Mat, University of Groningen,
Nijenborgh 4, NL-9747 AG Groningen, Netherlands.}
\footnotetext[3]{Groningen Cognitive Systems and Materials Center (CogniGron), University of Groningen,
Nijenborgh 4, NL-9747 AG Groningen, Netherlands}

\footnotetext[1]{Corresponding author}

\begin{document}
{
\pagestyle{empty}
\maketitle
}
\pagestyle{fancy}
\setcounter{footnote}{0}

\begin{abstract}
\begin{center}

Event-based vision sensors encode local pixel-wise brightness changes in streams of events rather than full image frames
and yield sparse, energy-efficient encodings of scenes, in addition to low latency, high dynamic range, and lack of motion blur.
Recent progress in object recognition from event-based sensors has come from conversions of
successful deep neural network architectures, which are trained with backpropagation.
However, using these approaches for event streams requires a transformation to a synchronous paradigm, which not only loses computational efficiency,
but also misses opportunities to extract spatio-temporal features.
In this article we propose a hybrid architecture for end-to-end training of deep neural networks for event-based pattern recognition and object detection,
combining a spiking neural network (SNN) backbone for efficient event-based feature extraction, and a subsequent
classical analog neural network (ANN) head to solve synchronous classification and detection tasks.
This is achieved by combining standard backpropagation with surrogate gradient training to propagate gradients
inside the SNN layers.
Hybrid SNN-ANNs can be trained without additional conversion steps, and result in highly accurate networks
that are substantially more computationally efficient than their ANN counterparts.
We demonstrate results on event-based classification and object detection datasets, in which only the architecture of
the ANN heads need to be adapted to the tasks, and no conversion of the event-based input is necessary.
Since ANNs and SNNs require different hardware paradigms to maximize their efficiency, we envision that SNN backbone and ANN head can
be executed on different processing units, and thus analyze the necessary bandwidth to communicate between the two parts.
Hybrid networks are promising architectures to further advance machine learning approaches for event-based vision,
without having to compromise on efficiency.

\end{center}
\end{abstract}

\clearpage

\section{Introduction}
\label{sec:intro}

\begin{figure}[t]
\begin{center}
   \includegraphics[width=0.69\linewidth]{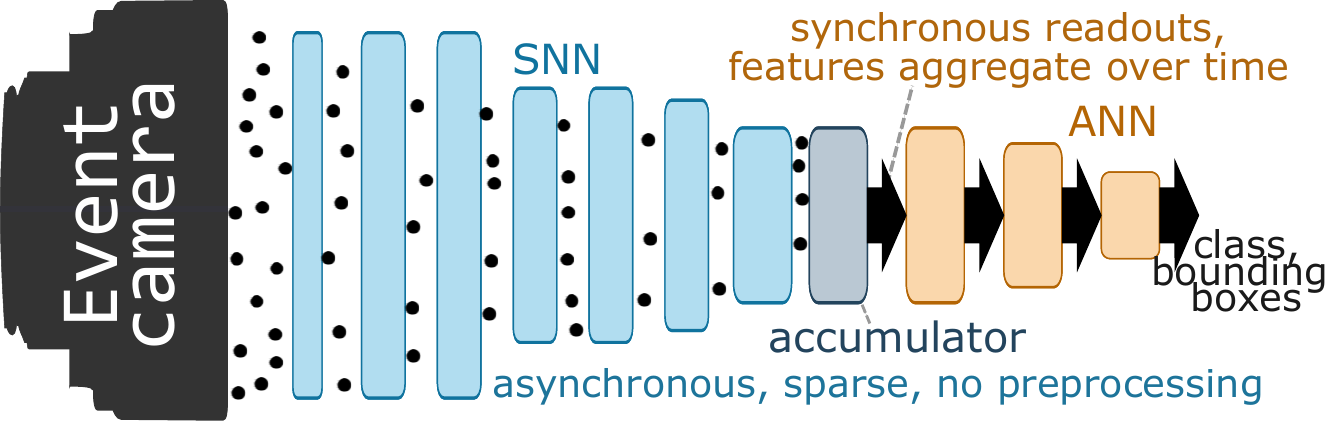}
\end{center}
   \caption{
Hybrid SNN-ANN models consist of an SNN backbone to compute features directly from event camera outputs which are accumulated and processed by an ANN head for classification and object detection.
Using sparse, binary communication (dots) instead of dense tensors (arrows) to process events enables efficient inference.
SNN and ANN can be on completely different devices.
}
\label{fig:method}
\end{figure}

Event-based vision sensors address the increasing need for fast and energy-efficient visual perception
\cite{Lichtsteiner_2008_566,posch_2011_259,liu2010neuromorphic,serrano_2013_827,Gallego_2020},
and have enabled new use cases such as high-speed navigation \cite{falanga_dynamic_2020},
gesture recognition \cite{maro2020event}, visual odometry and SLAM \cite{mueggler_2017_shapes,vidal_ultimate_2018,zhu2019neuromorphic}.
These sensors excel at very low latency and high dynamic range, while
their event-based encoding creates a sparse spatio-temporal representation of dynamic scenes.
Every event indicates the position, precise time, and polarity of a local brightness change.

In conventional frame-based computer vision deep learning-based methods have led to vastly improved
performance in object classification and detection, so it is natural to expect a boost
in performance also from applying deep neural networks to event-based vision.
However, such an approach has to overcome the incompatibility of machine learning algorithms
developed for a synchronous processing paradigm, and the sparse, asynchronous nature of event-based inputs.
Recent successful approaches for processing event data have therefore relied on early conversions of events into
filtered representations that are more suitable to apply standard machine learning methods \cite{amir2017low,Sironi_2018_CVPR,NEURIPS2020_perot}.
Biologically inspired spiking neural networks (SNNs) in principle do not require any conversion of event data and can process data
from event-based sensors with minimal preprocessing.

However, high performing SNNs rely on conversion from standard deep networks \cite{rueckauer2017,abhronil_2019},
thereby losing the opportunity to work directly with precisely timed events.
Other approaches like evolutionary methods \cite{vazquez_2019} or local learning rules like STDP \cite{kheradpisheh2018stdp,barbier2020unsupervised}
are not yet competitive in performance.

Here we describe a hybrid approach that allows end-to-end training of neural networks for event-based object recognition and detection.
It combines sparse spike-based processing of events in early layers with off-the-shelf ANN layers to process the sparse, abstract features (see \cref{fig:method} for an illustration).
This is made possible by combining standard backpropagation with recently developed surrogate gradient methods
to train deep SNNs \cite{neftci_2019_36,Lee2016_508,rathi2020dietsnn,shrestha2018,Wu_2019,cramer2020training}.
The advantage of the hybrid approach is that early layers can operate in the extremely efficient
event-based computing paradigm, which can run on special purpose hardware implementations \cite{Schemmel_2010_1947,furber2013_2454,Merolla_2014_668,qiao_2015_141,Davies_2018_82,billaudelle2019versatile}.
The hybrid approach is also optimized for running SNN parts and conventional neural networks on separate
pieces of hardware by minimizing the necessary bandwidth for communication.
In our experiments we demonstrate that the hybrid SNN-ANN approach yields very competitive accuracy
at significantly reduced computational costs, and is thus ideally suited for embedded perception applications
that can exploit the advantages of the event-based sensor frontend.

Our main contributions are as follows:
\begin{itemize}
\item We propose a novel hybrid architecture that efficiently integrates information over time without needing to transform input events into other representations.
\item We propose the first truly end-to-end training scheme for hybrid SNN-ANN architectures on event camera datasets for classification and object detection.
\item We investigate how to reduce communication bandwidths for efficient hardware implementations that run SNNs and ANNs on separate chips.
\item We analyze how transfer learning of SNN layers increases the accuracy of our proposed architecture.
\end{itemize}

\section{Related Work}
\label{sec:rel-work}

A variety of \textbf{low level representations} for event-based vision data have been explored:
The HOTS method \cite{lagorce_2017} defines a time surface,
\ie a two-dimensional representation of an event stream by convolving a kernel over the event stream.
This method was improved in \cite{Sironi_2018_CVPR} by choosing an exponentially decaying kernel and adding local memory for increased efficiency.
In \cite{Gehrig_2019_ICCV}, a more general take on event stream processing is proposed,
utilizing general kernel functions that can be learned and project the event stream to different representations.
Notably, using a kernel on event camera data and aggregating the result is the same as using a spiking neural network layer.
Our approach allows learning a more general low level representation by using a deep SNN with an exponentially decaying kernel,
compared to only one layer in \cite{Gehrig_2019_ICCV} and learnable weights compared to \cite{lagorce_2017,Sironi_2018_CVPR}.

\textbf{Conversion approaches} such as \cite{rueckauer2017,abhronil_2019} transform trained ANNs into SNNs for inference, and so far
have set accuracy benchmarks for deep SNNs.
Conversion methods train on image frames and do not utilize the membrane potential or delays to integrate information over time.
In \cite{kugele_2020_effproc} networks are unrolled over multiple time steps, which allows training on event camera datasets.
However, temporal integration is only encoded in the structure of the underlying ANN, but not in the dynamics of spiking neurons.
In their formulation, the efficiency of the SNN is limited by the rate coding of the neurons, which is potentially
more inefficient than encodings learned via end-to-end training in our hybrid SNN-ANN approach.
In addition, conversion methods typically do not optimize for energy-efficiency.

\textbf{SNN training with variants of backpropagation} has been demonstrated in \cite{Lee2016_508,shrestha2018,Wu_2019},
albeit on simpler architectures (\eg only one fully-connected layer in \cite{Lee2016_508}) and, in general, without delays during simulation.
A mixed approach is used in \cite{rathi2020dietsnn}, first training and converting an ANN and then training the converted SNN.
Our approach uses synaptic delays and skip connections, exploring how complex ANN architectures translate to SNN architectures.
The closest architecture to ours is from \cite{lee_2020_spikeflow}, which is trained from scratch to predict optical flow.
Their U-Net with an SNN encoder transmits information from all SNN layers, leading to a high bandwidth from SNN to ANN.
We improve on this by using only a single layer to transmit information from SNN to ANN, and extend to applications
in classification and object detection tasks.

\textbf{Conventional ANN architectures} are used in \cite{Rebecq_2019_CVPR} to solve the task of image reconstruction from events with a recurrent U-Net,
and they subsequently show that classification and object detection are possible on the reconstructed images.
No SNN layers are used in this case, resulting in a computationally expensive network and the need to preprocess the event camera data.
Faster and more efficient training and inference for object detection is presented in \cite{NEURIPS2020_perot},
who propose a recurrent convolutional architecture that does not need to explicitly reconstruct frames.
As the sparsity of the event stream is not utilized, it is expected that gains in energy-efficiency are possible with SNN approaches.

\section{Methods}
\label{sec:methods}

\begin{figure}[t]
\begin{center}
   \includegraphics[width=0.999\linewidth]{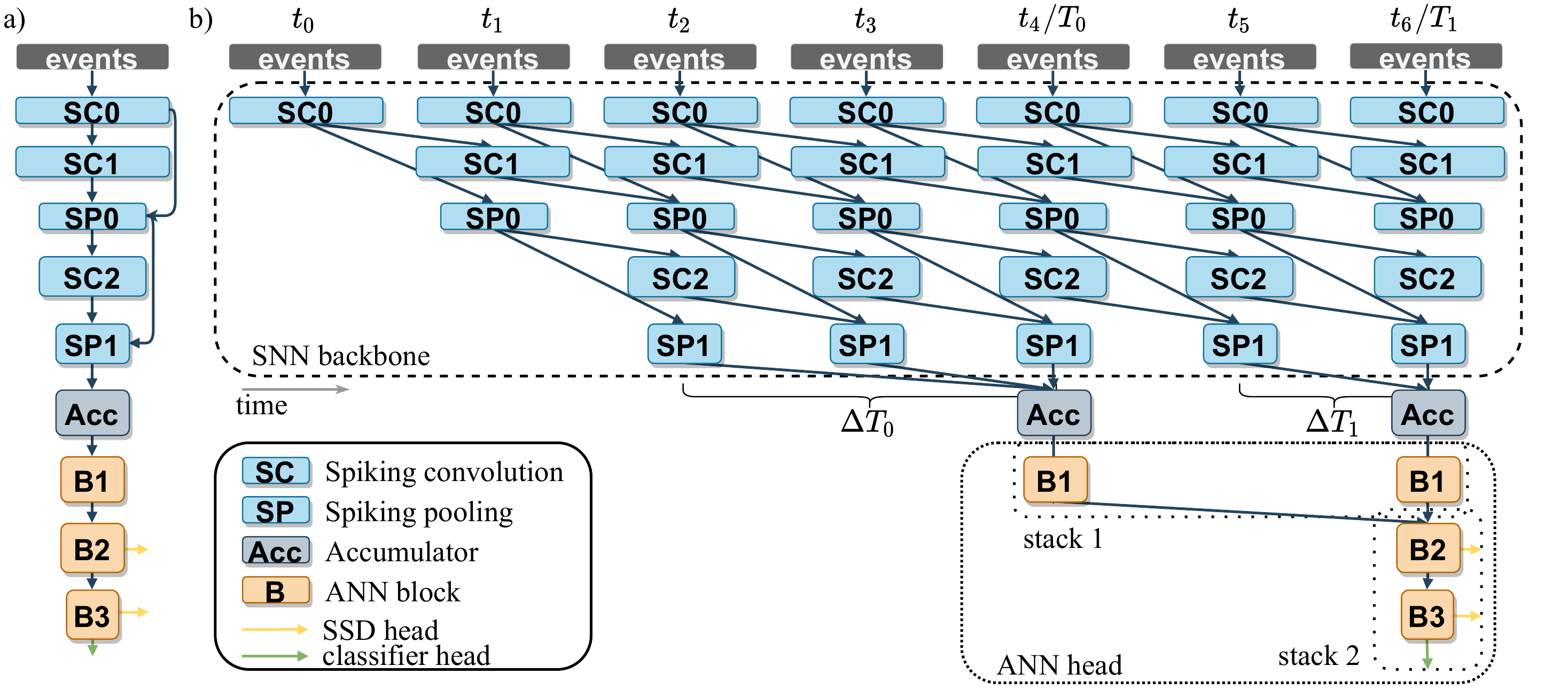}
\end{center}
   \caption{
Training a hybrid network with a DenseNet backbone.
a) Compact representation.
b) Network rolled out over time steps.
The SNN backbone computes sparse features from event camera data that are accumulated at time intervals $\outint_i$.
The ANN head integrates features from multiple outputs (2, in this example) for a prediction (classification or object detection).
We use a time step of \SI{1}{ms} to integrate information over small time scales.
During inference, the SNN backbone runs asynchronously without a time-stepped rollout, enabling potential savings in computation.
Layers with the same name share weights.
}
\label{fig:hybridnet}
\end{figure}

This section introduces the proposed hybrid SNN-ANN architecture and describes training, inference and metrics we used to evaluate our method.
Our hybrid network consists of two parts: An SNN backbone that computes features from raw event camera inputs,
and an ANN head that infers labels or bounding boxes at predefined times (see \cref{fig:hybridnet}).
The overall task is to find an efficient mapping
from a sequence of event camera data $E$
in a time interval $T$ to a prediction $P$, which can be a label $l$ or a set of bounding boxes $B$.
Our approach consists of three stages:
First, continuously in time, an intermediate representation $I = S(E)$ is generated using the SNN backbone.
Second, this intermediate representation is accumulated at predefined points in time.
Third, when all accumulators are filled, the accumulated intermediate representations are mapped via the ANN head $A$ to the final prediction $P = A(\accum(I)) = A(\accum(S(E)))$.
The following sections describe all three parts in more detail.

\subsection{SNN backbone}

Spiking neural networks (SNNs) are biologically inspired neural networks, where each neuron has an internal state.
Neurons communicate via spikes, i.e.\ binary events in time, to signal the crossing of their firing thresholds.
Upon receiving a spike $i$, the synaptic current $\current$ changes proportionally to the synaptic weight $\synweight$, which in turn leads to a change of the neuron's internal state $\state$.
Because of the binary and sparse communication, these networks can be significantly more energy-efficient than dense matrix multiplications in conventional ANNs (see also \cite{rieke1999spikes}).

The task of the SNN backbone $S$ is to map a sequence of raw event camera inputs $E$ into a
compressed, sparse, and abstract intermediate representation  $I = S(E)$ in an energy-efficient way.
More concretely, the input is a stream of events $e_i = (t_i, x_i, y_i, p_i)$, representing the time $t_i$ and polarity $p_i$ of an input event at location  $(x_i, y_i)$.
Polarity is a binary $\{-1, 1\}$ signal that encodes if the brightness change is positive (brighter) or negative (darker).
This stream is processed by the SNN without further preprocessing.
In our implementation, the first layer has two input channels representing the two polarities.
In contrast to previous work such as \cite{lee_2020_spikeflow} and \cite{kugele_2020_effproc},
the input events are never transformed into voxel grids or rate averages,
but directly processed by the SNN to compute the intermediate representation $S(E)$.
The spiking neuron model we use is the leaky integrate-and-fire model \cite{gerstner_2014_neurondyn}, simulated using the forward Euler method,
\begin{align}
\current_i &= \synleak \current_{i-1} + \sumsmall_j w_{j}\spikes_j\\
\label{eq:neuronstate}
\state_i &= \memleak \state_{i-1} + \current_i\\
\spikes_i &= \Theta(\state_i - \thresh)
\label{eq:spikes}
\end{align}
where $\state$ is the membrane potential, $\current$ is the presynaptic potential, $\spikes_{j,i}$ are the binary input and output spikes, $\thresh$ is the threshold voltage, and $\timestep = t_i - t_{i-1}$ is an update step.
The membrane leakage $\memleak$ and the synaptic leakage $\synleak$ are given as $\exp{(-\timestep / \tau_\mathrm{mem})}$ and $\exp{(-\timestep / \tau_\mathrm{syn})}$, respectively.
To simulate the membrane reset, we add a term to \cref{eq:neuronstate} which implements the reset-by-subtraction \cite{rueckauer2017} mechanism,
\begin{equation}
\state_i = \state_i - \thresh \spikes_{i-1} ~~~.
\end{equation}
The threshold is always initialized as $\thresh=1$ and trained jointly with the weights (see appendix for details).

We simulate our network by unrolling it in time with a fixed time step $\timestep$.
\Cref{fig:hybridnet}b shows the training graph for a rollout of 7 time steps.
Our simulation allows choosing arbitrary delays for the connections of different layers, which determines how information is processed in time.
Inspired by recent advances, we choose to implement streaming rollouts \cite{fischer2018} in all our simulations.
This means that each connection has a delay of one time step, in accordance with the minimum delay of large-scale simulators for neuroscience \cite{Stimberg2019}.
This allows integrating temporal information via delayed connections, in addition to the internal state.

The SNN backbone $S(E)$ outputs a set of sequences of spikes in predefined time intervals $\tout$: $I = (e_j)_{t_j \in \tout}$.
An example is shown in the dashed box of \cref{fig:hybridnet}.
Details about the backbones used can be found in \cref{sec:backbones}.
The figure shows an unrolled DenseNet backbone with two blocks (SC0 to SP0 and SP0 to SP1, respectively), where two output intervals are defined as $\outint_0 = [t_2, T_0 ]$ and $\outint_1 = [t_5, T_1]$.
This structure is used during training, where a time-stepped simulator approximates the continuous-time SNN.
During inference, the SNN backbone runs asynchronously, enabling savings in computation.

\subsection{Accumulator}

Our model connects the sparse, continuous representation of the SNN with the dense,
time-stepped input of the ANN with an accumulator layer for each output interval $\outint_i$ (Acc in \cref{fig:hybridnet}).
The task of this layer is to transform the sparse data to a dense tensor.
We choose the simple approach of summing all spikes in each time interval $\outint_i$ to get a dense tensor with the feature map shape $(\nchannels, \fmapheight, \fmapwidth)$.

\subsection{ANN head}

The ANN head processes the accumulated representations from the accumulators to predict classes or bounding boxes.
The general structure of the ANN head can be described with three parameters: the number of SNN outputs $\nouts$, the number of stacks $\nstacks$ and the number of blocks per stack $\nblockstack$.
The exemplary graph in \cref{fig:hybridnet} has $\nouts=2$, $\nstacks=2$ and $\nblockstack=2$.
Having multiple outputs and stacks allows to increase the temporal receptive field, \ie the time interval the ANN uses for its predictions (for details see appendix).
All blocks with the same name share their weights to reduce the number of parameters.
The number of blocks can be different for each stack.
In most experiments we use two stacks with 1 and 3 blocks, respectively.
The dense representation for each $\outint_i$ is then further processed by each stack, where results are summed at the end of a stack and used as input for the next stack.
Each block in a stack consists of batch normalization $\batchnorm$, convolution $\conv$, dropout $\dropout$, $\relu$ and pooling $\pool$,
\eg $\mathrm{B0}(x) = \pool(\relu(\dropout(\conv(\batchnorm(x)))))$.
We use a convolutional layer with kernel size 2, stride 2, learnable weights and $\relu$ activation as pooling layer.
Whenever we use dropout in conjunction with weight sharing, we also share the dropout mask.
In the case of classification, a linear layer is attached to the last block in the last stack (see \cref{sec:class}).
For object detection, an SSD head is attached to selected blocks in the last stack (see \cref{sec:od}).

\subsection{Energy-efficiency}
\label{sec:energy}
We use the same metric as \cite{rueckauer2017}, \ie we count the number of operations of our network.
Due to the sparse processing and binary activations in the SNN, the number of operations is given
by the synaptic operations, \ie the sum of all postsynaptic current updates.
For ANN layers, we count the number of multiply-add operations, as the information is processed with dense arrays.
For this metric, it is assumed that both SNN and ANN are run on dedicated hardware.
Then the total energy is proportional to the number of operations plus a constant offset.
We discuss benefits, drawbacks and alternatives to this metric in the appendix.
To regularize the number of spikes we utilize an $\mathrm{L}_1$ loss on all activations of the SNN backbone
\begin{equation}
L_\mathrm{s} =  \sum_{l,b,i,x,y}\frac{\lambdaspikes}{LBTW_lH_l} |S_{l,i,b,x,y}| .
\end{equation}
with a scaling factor $\lambdaspikes$, the number of layers $L$, the batch size $B$, total simulation time $T$,
and width $W_l$ and height $H_l$ of the respective layer.
This also reduces the bandwidth, as discussed in \cref{sec:bandwidth}.

\subsection{Bandwidth}
\label{sec:bandwidth}
As we expect that special hardware could be used to execute at least the SNN backbone during inference,
we want to minimize the bandwidth between SNN and ANN to avoid latency issues.
We design our architectures such that only the last layer is connected to the ANN and use $\mathrm{L}_1$ loss
on the activations to regularize the number of spikes.
This is in contrast to \cite{lee_2020_spikeflow}, where each layer has to be propagated to the ANN.
We report all bandwidths in MegaEvents per second (\si{MEv/s}) and MegaBytes per second (\si{MB/s}).
One event equals \SI{6.125}{Bytes},
because we assume it consists of 32 bit time + 16 bit spatial coordinates + 1 bit polarity.

\subsection{Classification}
\label{sec:class}
To classify, we attach a single linear layer to the last block in the last stack of our hybrid network.
We use the negative log-likelihood loss of the output of this layer.
Additionally, we use $\mathrm{L_2}$-regularization (weight decay) with a factor of $0.001$.
We use the Adam optimizer \cite{Kingma2015AdamAM} with a learning rate of $\lr=0.01$
and a learning rate schedule to divide it by 5 at \SI{20}{\%}, \SI{80}{\%} and \SI{90}{\%} of the total number of epochs $\numeps=100$.

\subsection{Object detection}
\label{sec:od}
We use the SSD (single shot detection) architecture \cite{Liu_2016}, which consists of a backbone feature extractor and multiple predictor heads.
Features are extracted from the backbone at different scales and for each set of features,
bounding boxes and associated classes are predicted, relative to predefined prior boxes.
In the appendix, we present a novel, general way to tune the default prior boxes in a fast and efficient way.
During inference, non-maximum suppression is used on the output of all blocks to select non-overlapping bounding boxes with a high confidence.
The performance of the network is measured with the VOC mean average precision (mAP) \cite{everingham2009the}.
If not otherwise denoted, we use the same learning hyperparameters as in \cref{sec:class} but a smaller learning rate of $0.001$.

\subsection{Backbone architectures}
\label{sec:backbones}

\begin{figure}[t]
\begin{center}
   \includegraphics[width=0.89\linewidth]{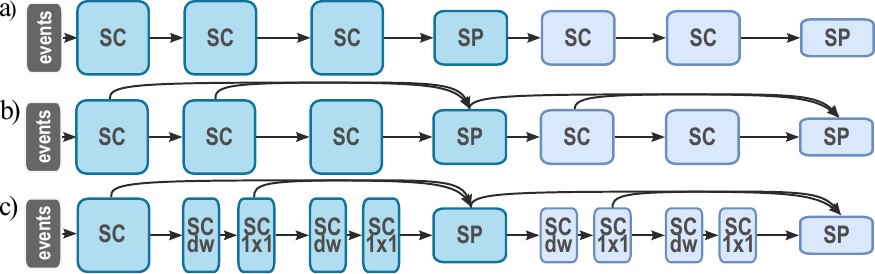}
\end{center}
   \caption{
The different backbones in compact representation with spiking convolutional (\textbf{SC}) and pooling (\textbf{SP}) layers.
a) VGG.
b) DenseNet.
c) \DenseSep{} (DenseNet with depthwise separable convolutions).
Depthwise separable convolutions consist of a depthwise convolution (\textbf{dw}) and a convolution with kernel size \textbf{1x1}.
Shades of blue mark different blocks.
All depicted networks have 2 blocks and 2 layers per block.
Multiple inputs are concatenated.
}
\label{fig:backbones}
\end{figure}

Three different architecture types are used in our experiments:
VGG \cite{simonyan2015vgg}, DenseNet \cite{Huang_2016} and \DenseSep.
A VGG network with $\nblocks$ blocks and $\nlayersperblock$ layers per block is a feed-forward neural network
with $\nblocks\cdot\nlayersperblock$ layers, where each output channel
is given by $\growthfactor \cdot l$ with $l$ the layer index (starting from 1).
The DenseNet structure consists of $\nblocks$ blocks,
where each block consists of $\nlayersperblock$ connected layers per block.
\DenseSep{} is a combination of the depthwise separable block of MobileNet \cite{howard2017mobilenets}
and the DenseNet structure (see \cref{fig:backbones}).

\section{Results}
\label{sec:results}

\subsection{Classification on \nmnist{}}
\label{sec:nmnist}

\begin{figure}[t]
\begin{center}
   \includegraphics[width=0.494\linewidth]{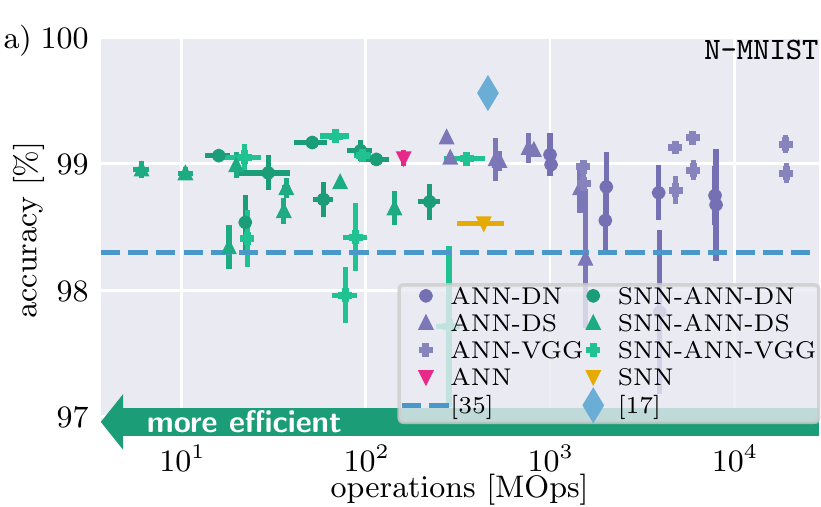}
   \includegraphics[width=0.494\linewidth]{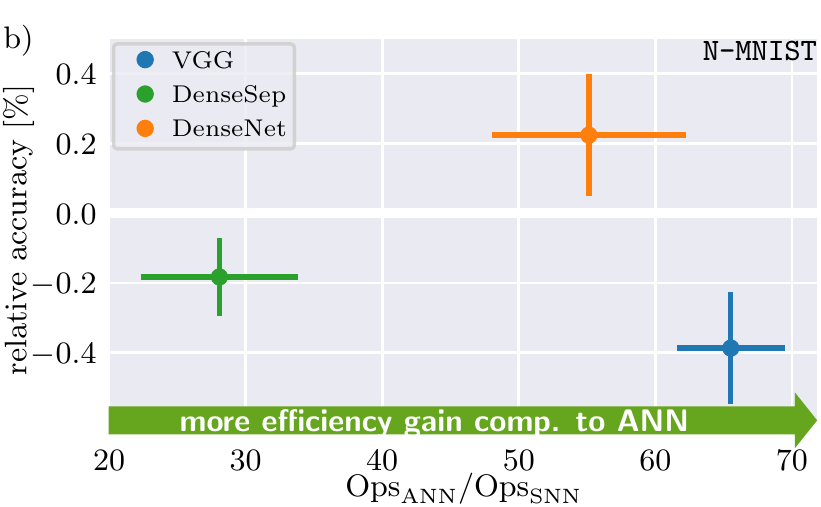}
\end{center}
   \caption{
\textbf{a)} Accuracy on the \nmnist{} test set vs.\ number of operations for different architectures.
Hybrid SNN-ANN architectures (green) overcome the efficiency limit of conversion-based architectures (blue diamond) with only a minor drop in accuracy.
SNN-ANN architectures are more efficient than almost all ANNs (purple).
Compared to the SNN and ANN baseline, our hybrid networks increase accuracy and energy-efficiency.
\textbf{b)} Relative accuracy on the \nmnist{} test set vs.\ relative number of operations.
The VGG backbone has the highest gain in energy-efficiency, while losing significantly in accuracy.
Other backbones show a minor decrease with a significant gain in energy-efficiency.
We report the mean and error of the mean over 6 repetitions.
}
\label{fig:AccOpsNmnist}
\label{fig:AccOpsNmnistRel}
\end{figure}

We train a hybrid SNN-ANN network for classification on the \nmnist{} dataset \cite{Orchard_2015_437}.
In \nmnist{}, each \mnist{} digit is displayed on an LCD monitor while an event camera performs three saccades within roughly $\SI{300}{ms}$.
It contains the same \num{60000} train and \num{10000} test samples as \mnist{} at a resolution of $34\times34$ pixels.
Here we compare the performance of our hybrid network to networks where the SNN backbone
is replaced with ANN layers (ANN-ANN), two baselines and 
two approaches from the literature: A conversion approach \cite{kugele_2020_effproc}, where a trained ANN
is converted to a rate-coded SNN
and an ANN that reconstructs frames from events and classifies the frames \cite{Rebecq_2019_CVPR}.
The first baseline is a feedforward ANN with the same structure as our VGG SNN-ANNs,
but where all time steps are concatenated and presented to the network as different channels.
The second baseline is an SNN of the same structure, where we accumulate the spikes of the last linear layer
and treat this sum as logits for classification.
The SNN is unrolled over the same number of time steps than the SNN-ANNs.
We report results for multiple network sizes and configurations 
to show that hybrid SNN-ANNs generally perform better in terms of energy-efficiency and bandwidth
(detailed architecture parameters can be found in the appendix).
Mean values of accuracy and number of operations are reported together with the error of the mean over 6 repetitions,
using different initial seeds for the weights.
For one \DenseSep{} network only 5 iterations are reported,
because training of one trial did not converge,
resulting in a significant outlier compared to the performance of all other networks.
In \cref{fig:AccOpsNmnist}a, we show the accuracy on the \nmnist{} test set vs.~the number of operations
for our architectures, baselines and related approaches from the literature.
Our hybrid networks (green, SNN) reach similar accuracies to the ANNs (purple), while being consistently more energy-efficient.
The best hybrid architecture is a DenseNet with $\growthfactor = 16$, $\nlayersperblock = 3$ and $\nouts = 1$.
Compared to \cite{kugele_2020_effproc},
it performs slightly worse ($\SI{99.10\pm0.09}{\%}$ vs.\ $\SI{99.56\pm0.01}{\%}$) in accuracy,
but improves significantly on the number of operations ($\SI{94\pm17}{MOps}$ vs.\ $\SI{460\pm38}{MOps}$)
despite having more parameters ($\num{504025}$ vs.\ $\num{319890}$).
The average bandwidth between SNN and ANN is $\SI{0.250\pm0.044}{MEv/s}$ for this architecture,
or approximately \SI{1.53}{MB/s}.
Our smallest DenseNet with $~\num{64875}$ parameters
($\growthfactor=8$, $\nlayersperblock=2$, $\nouts=1$) is even more efficient,
needing only $\SI{15.9\pm2.5}{MOps}$ at a bandwidth of $\SI{0.144\pm0.034}{MEv/s}$
to reach $\SI{99.06\pm0.03}{\%}$ accuracy.
Our results are mostly above \cite{Rebecq_2019_CVPR} in terms of accuracy, although our networks are much smaller
(their networks have over \SI{10}{M} parameters).
Due to the large parameter size, we also estimate that we should be more energy-efficient, although the authors do
not provide any numbers in their publication.

In \cref{fig:AccOpsNmnistRel}b, we plot the average per backbone over all architectures in \cref{fig:AccOpsNmnist}a,
relative to the averages of the ANN-ANN architectures.
All hybrid SNN-ANNs improve the number of operations significantly over ANN-ANN implementations
with at most a minor loss in accuracy.
Hybrid DenseNets provide the best accuracy-efficiency trade-off with an average improvement of about a factor of 56
in energy-efficiency while also increasing accuracy by approximately \SI{0.2}{\%}.
For VGG architectures, the energy-efficiency is increased by a factor of roughly 65 at a loss in accuracy of 0.4
between hybrid and ANN-ANN architectures.
Hybrid \DenseSep{} architectures lose approximately 0.2 accuracy points,
but gain a factor of about 28 in energy-efficiency.
We assume that the \DenseSep{} architectures perform the worst in comparison for two reasons:
First, they are already the most efficient ANN architectures,
so improving is harder than for less optimized architectures.
Second, as the effective number of layers is higher in this architecture compared to the other two,
optimizing becomes more difficult and gradient deficiencies
(vanishing, exploding, errors of surrogate gradient) accumulate more.
Hyperparameter optimization (learning rate, dropout rates, regularization factors for weights and activations)
was not performed for each network separately,
but only once for a DenseNet with $\nlayersperblock=2, \growthfactor=16, \nouts=1$
on a validation set that consisted of \SI{10}{\%} of the training data, \ie 6000 samples.

\subsection{Classification on \ncars{}}
\label{sec:ncars}

\ncars{} is a binary car vs.\ background classification task, where each sample has a length of \SI{100}{ms}.
We train the same networks as in \cref{sec:nmnist} with a growth factor $\growthfactor=16$ and compare to the same
baselines and two results from the literature.
We see the same trend in energy-efficiency improvements over ANNs,
with factors ranging from 15 (\DenseSep{}) to 110 (VGG).
Relative accuracy decreases significantly for \DenseSep s by 1.5 points, but is 0.75 points higher for VGG.
These results, together with the results in \cref{fig:AccOpsNmnistRel}b suggest, that the more efficient an architecture
already is, the less can be gained from training an equivalent hybrid SNN, although the gains of at least a factor of
15 in energy-efficiency is still significant.
In comparison to our ANN and SNN baselines, our SNN-ANNs perform better in terms of accuracy and number of operations.
Two out of four SNN runs could not go beyond chance level and were therefore excluded from the validation.
In conclusion, using hybrid SNN-ANNs saves energy compared to SNNs and ANNs of similar structure.
In this experiment, our architecture is not competitive to state-of-the-art \cite{kugele_2020_effproc} in accuracy,
but has a similar energy demand with approximately double the number of parameters.
Detailed results and figures can be found in the appendix.

\subsection{Object detection on \shapes{}}
\label{sec:shapesOD}

The \shapes{} dataset \cite{mueggler_2017_shapes} contains frames, events,
 IMU measurements and calibration information of a scene of different shapes pinned to a wall.
As training data for event-based vision object detection is scarce \cite{Gehrig_2020_CVPR},
we labelled bounding boxes for each of the \num{1356} images and \num{10} shapes,
resulting in \num{9707} ground truth bounding boxes.
A detailed description and an example of the dataset can be found in the appendix.
We provide two train/test splits:
\shapesEasy{} where \SI{90}{\%} of the data is randomly assigned to the train set
and a more difficult split \shapesHard{},
where only \SI{30}{\%} of the data is used in the train set.
In section \cref{sec:shapesEasy} and \cref{sec:shapesHard}, we present the results on \shapesEasyHard{}.

\subsection{Results on \shapesEasy{}}
\label{sec:shapesEasy}

\begin{figure}[t]
\begin{center}
   \includegraphics[width=0.494\linewidth]{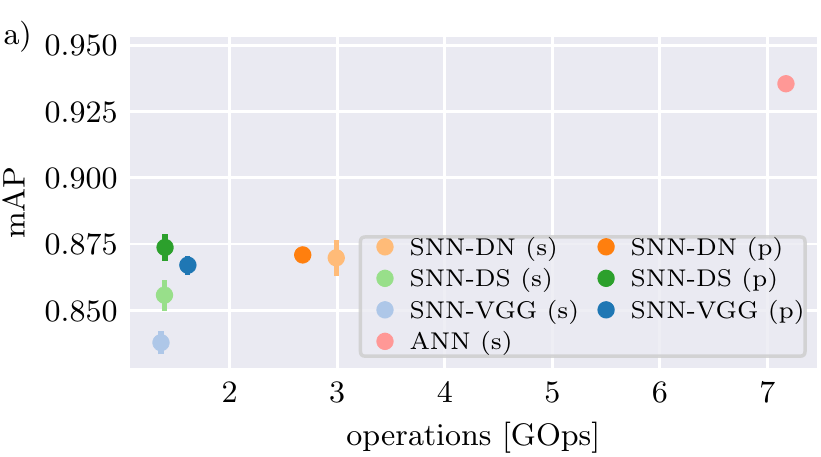}
   \includegraphics[width=0.494\linewidth]{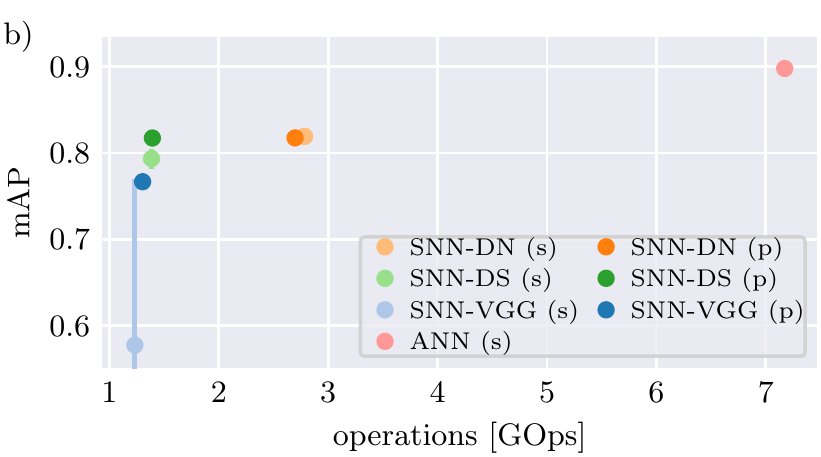}
\end{center}
   \caption{
Mean average precision (mAP) over number of operations for \shapesEasyHard{} (\textbf{a}/\textbf{b}, mean over 4 trials).
The architectures with the best results on the \nmnist{} training are either trained from scratch (s) or initialized with the weights from the training (pretrained, p).
Pretrained architectures improve over random initialization (VGG, \DenseSep{}) or are on par with it (DenseNet).
}
\label{fig:shapesEasy}
\label{fig:shapesHard}
\end{figure}

For all backbones, we take the architecture parameters from the best network in the \nmnist{} task.
See the appendix for details.
We want to compare hybrid networks trained from scratch
with networks where the SNN backbone is initialized with network weights trained on \nmnist{}.
This allows investigating the effect of transfer learning during training.
Results are shown in \cref{fig:shapesEasy}.
The networks with pretrained weights always converge to higher mAP than their non-pretrained counterpart.
The DenseNet and \DenseSep{} backbones perform better than VGG.
This is in agreement with the results for classical ANNs,
where DenseNet architectures outperform VGG on image-based datasets like ImageNet.
Our best network is a pretrained SNN-ANN \DenseSep{} with a mean average precision of $\SI{87.37\pm0.51}{\%}$,
$\SI{1398.9\pm2.3}{MOps}$ operations and a bandwidth of \SI{11.0}{MB/s}.
A comparable ANN backbone would require a bandwidth of \SI{1210}{MB/s}.
A regular SSD architecture with the same backbone as our VGG network,
where all time steps are concatenated over the channels outperforms our networks in terms of mAP,
but also needs significantly more operations.
We report the detailed results in the appendix.

\subsection{Results on \shapesHard{}}
\label{sec:shapesHard}

We do the same evaluation as in \cref{sec:shapesEasy},
but with a \SI{30/70}{\%} training/test data split (\cref{fig:shapesHard}).
The mAP is higher for networks with pretrained weights for \DenseSep{} and VGG and on par with DenseNet.
As with \shapesEasy{}, the backbones with skip connections perform better than the VGG backbone.
One of the four VGG experiments did not fully converge, explaining the high standard deviation.
Our best network is an SNN-ANN DenseNet trained from scratch with $\SI{2790\pm50}{MOps}$ operations,
a mean average precision of $\SI{82.0\pm1.0}{\%}$,
and a bandwidth of \SI{2.68}{MB/s}.
A comparable ANN backbone would have a bandwidth of \SI{864}{MB/s}.
As in \cref{sec:shapesEasy} the regular SSD architecture is better in mAP but worse in the number of operations.
We report the detailed results in the appendix.

\section{Conclusion}
\label{sec:discussion}

In this paper, we introduced a novel hybrid SNN-ANN architecture for efficient classification and object detection on event data that can be trained end-to-end with backpropagation.
Hybrid networks can overcome the energy-efficiency limit of rate-coded converted SNN architectures,
improving by up to a factor of 10.
In comparison to similar ANN architectures,
they improve by a factor of 15 to 110 on energy-efficiency with only a minor loss in accuracy.
Their flexible design allows efficient custom hardware implementations for both the SNN and ANN part, while minimizing the required
communication bandwidth.
Our SNN-ANN networks learn general features of event camera data, that can be utilized to boost the object detection performance on a transfer learning task.

We expect that the generality of the features can be improved when learning on larger and more diverse datasets.
Our work is particularly suited for datasets where temporal integration happens on a short time interval,
but struggles for longer time intervals, \eg multiple seconds due to the immense number of rollout steps needed.
Recent advances in deep learning, particularly C-RBP \cite{NEURIPS2020_drew} can potentially help to overcome this
by using recurrent backpropagation that has a constant memory complexity with number of steps (compared to backpropagation, where the memory-complexity is linear).
This also would ensure that our networks converge to a fixed point over time, potentially making predictions more stable.
More work on surrogate gradients and methods to stabilize training can further help to increase both energy-efficiency and performance of hybrid networks.
Our work is a first step towards ever more powerful event-based perception architectures that are going to challenge the performance of image-based deep learning methods.

\section*{Acknowledgments}
The authors would like to acknowledge the financial support of the CogniGron research center
and the Ubbo Emmius Funds (Univ. of Groningen).
Furthermore, this publication has received funding from the European Union’s Horizon 2020 research
innovation programme under grant agreement 732642 (ULPEC project).

{
\small
\bibliographystyle{unsrtnat}
\bibliography{egbib}
}

\clearpage
\appendix
\begin{center}
\huge\textbf{Appendix}
\end{center}

\section{Metrics for Energy-Efficiency}
We chose to use the metric of \cite{rueckauer2017} to compare the number of operations between ANNs and SNNs as a
metric to compare the energy-efficiency of these two types of neural networks.
This is a simplification, because in reality the actual energy consumption also depends on the specific hardware
that the networks run on.
However, this is a widely adopted metric because it is simple to calculate and allows for an approximate comparison
\cite{neil2016,rueckauer2017,bose2020neural,kugele_2020_effproc,davidson_2021}.
In \cite{thakur_2018}, an overview over different neuromorphic architectures is given
that shows a range of $2.8$-$360~\mathrm{pJ}$ per synaptic operation, \ie a factor of 100 between different
hardware implementations is possible for SNNs.
Also, this value is dependent on even more factors,
like networks size (how much of the hardware is used by the network)
and the base energy needed to operate the hardware.
Comparing actual energy measurements also suffers from the fact that you have to ensure to utilize the chips in the
same way, which is barely possible.
In \cite{rathi2020dietsnn}, a factor of approximately 5 is introduced to correct the differences in energy consumption
of an operation of SNN and ANN, which would significantly increase the energy-efficiency measured in our experiments
to factors of 140 to 325 between SNNs and ANNs.
Although our results would look more impressive, we don't think that this is a fair comparison based on the discussion
in this section.
In \cite{davidson_2021}, a more thorough analysis of the number of operations is provided that takes all operations
on digital hardware into account and shows that rate-coded SNNs are not more efficient than ANNs with floating point
activations.
As the number of operations of our networks is well below rate-coded SNNs, we also save energy compared to our ANNs.

\section{Classification on \ncars{}}
\label{sec:hybnetapp:ncars}

\begin{figure}[t]
\begin{center}
   \includegraphics[width=0.494\linewidth]{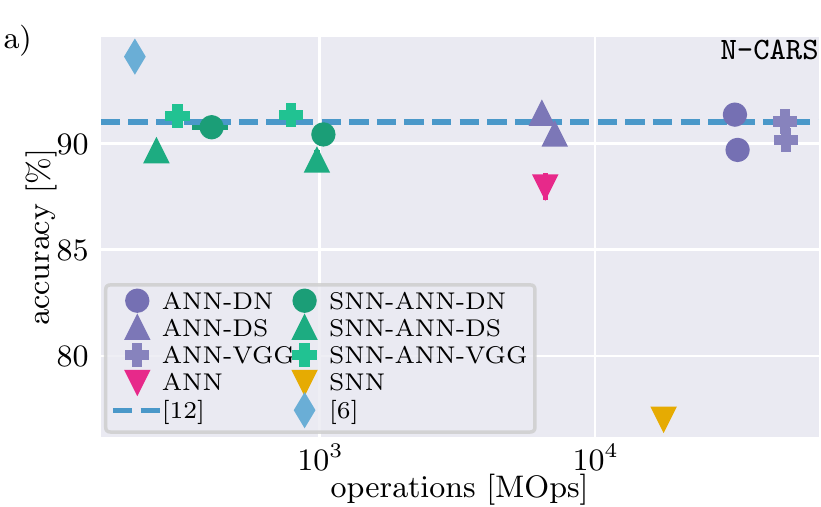}
   \includegraphics[width=0.494\linewidth]{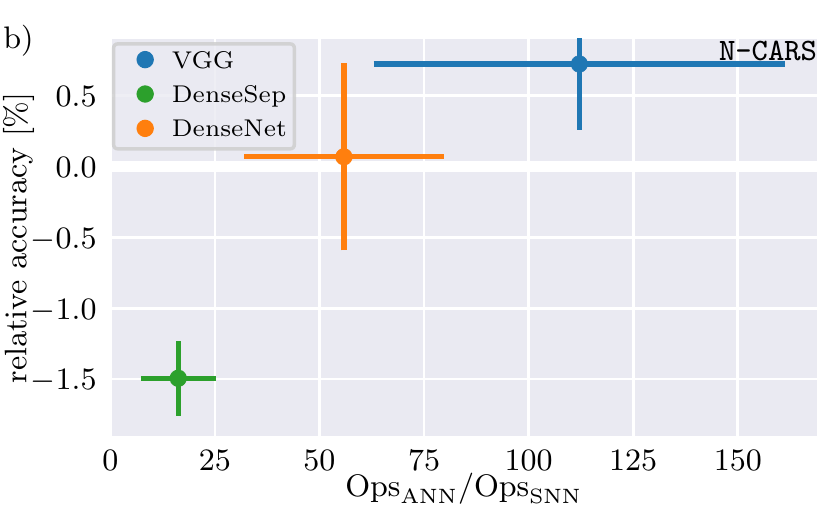}
\end{center}
   \caption{
\textbf{a)} Accuracy on the \ncars{} test set vs.\ number of operations for different architectures.
Hybrid SNN-ANN architectures are more efficient than all ANNs (purple), the most efficient ones are on par in 
efficiency with the conversion approach.
Compared to the SNN and ANN baseline, our hybrid networks increase accuracy and energy-efficiency.
\textbf{b)} Relative accuracy on the \ncars{} test set vs.\ relative number of operations for the three different backbones.
As with \nmnist{}, VGG backbones gain the most, followed by DenseNet and \DenseSep{}.
We report the mean and error of the mean over 4 repetitions.
}
\label{fig:hybnetapp:AccOpsNcars}
\label{fig:hybnetapp:AccOpsNcarsRel}
\end{figure}

We report the results of our experiments in \cref{fig:hybnetapp:AccOpsNcars}.
Our hybrid networks improve in energy-efficiency compared to our ANN-ANNs, as well as to the baselines.
Furthermore, our hybrid networks show a significantly higher accuracy compared to the baselines.
They are on par with \cite{Rebecq_2019_CVPR}, while using smaller networks.
They are, however, not yet competitive compared to the conversion approach of \cite{kugele_2020_effproc}.

\section{More information about backbones}
In this section, we list more details about the different backbones.
Each layer in a block is connected to all layers in the same block before it and has $\growthfactor$ output channels.
Multiple inputs are concatenated to preserve the temporal structure of the data.
To match the structure of the DenseNet, we employ the same hyperparameters also for the other architectures.
All pooling layers are depthwise convolutional layers with kernel size and stride 2,
because then the layer can adapt to the input activation during training.
With regular pooling layers, it can happen that the activation after the pooling layer dies out.
Compared to the DenseNet architecture,
adding depthwise separable convolutions together with 1x1 convolutions
should reduce the number of operations of the ANN significantly.
We want to evaluate if this still holds when using these building blocks in our SNN backbone.

\section{Surrogate gradients}
\label{sec:surrogates}

\begin{figure}[t]
\begin{center}
   \includegraphics[width=0.7\linewidth]{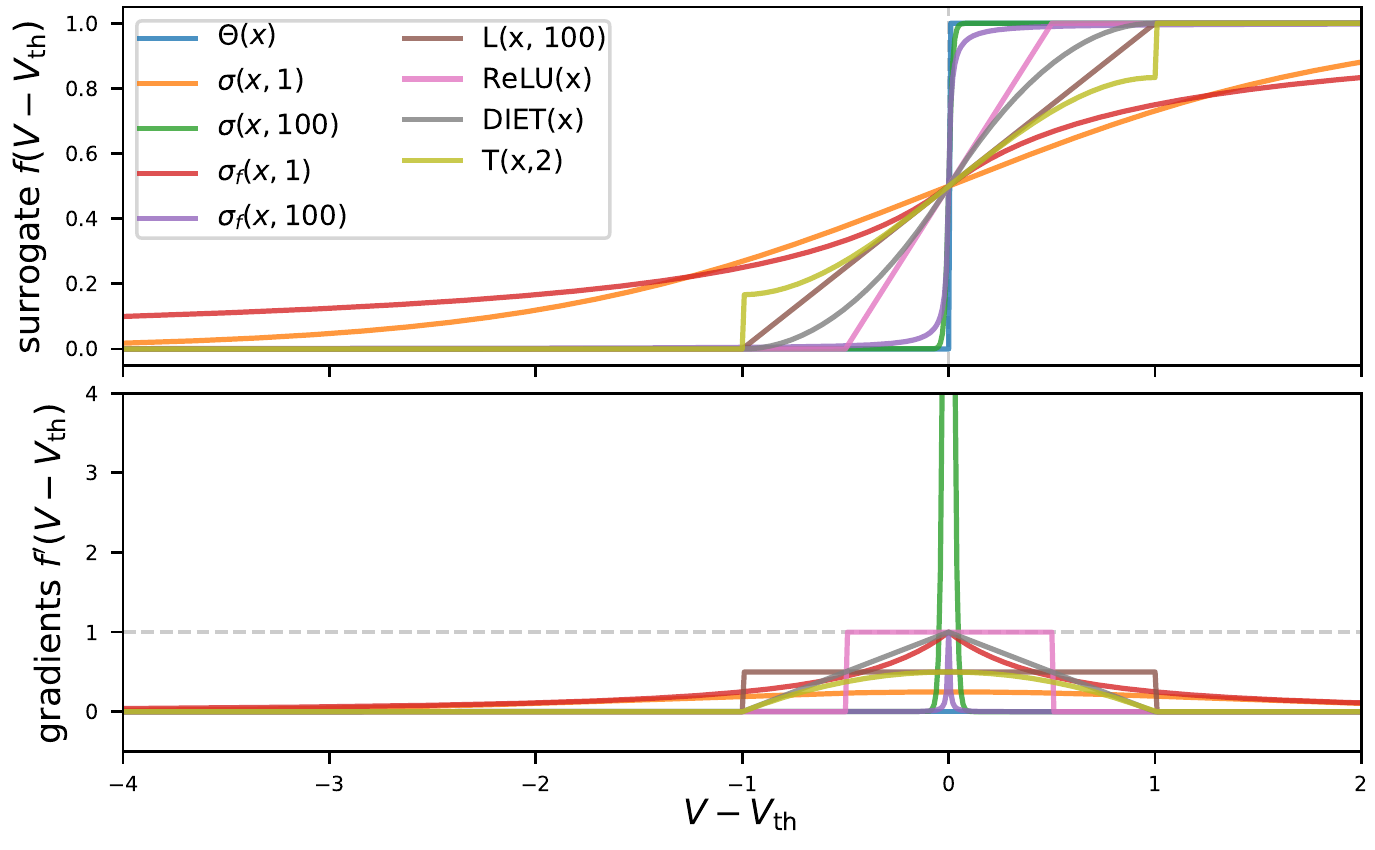}
\end{center}
   \caption{
Surrogate functions ({\it top}) and their corresponding analytical gradients ({\it right}).
A tradeoff between accurately approximating the function $\Theta(V-V_\mathrm{th})$ and allowing non-zero gradients has to be made.
The range of the x-axis has been chosen to be approximately in the range of typical values in our experiments and $\thresh=1$.
}
\label{fig:hybnetapp:surrogate}
\end{figure}

Surrogate gradients are used to replace the gradient of the activation function $\frac{\partial\Theta(x)}{\partial x}$ of the spiking network in \cref{eq:spikes},
because the gradient of this function is 0 everywhere, except for at 0, where it is ill-defined.
Recently, multiple surrogate gradients have been proposed.
The goal is to find a function that has a well-defined gradient,
ideally close to 1 to avoid vanishing and exploding gradient issues, while the approximate function is as close as possible to the activation function \cref{eq:spikes}.

A natural approach is to replace the $\Theta$-function with a sigmoid $\sigma$ plus an additional parameter $a$ that controls the steepness (orange and green line in \cref{fig:hybnetapp:surrogate})
\begin{equation}
\sigma(x,a) = \frac{1}{1+\exp{(-ax)}}.
\end{equation}
The steepness then controls the tradeoff between the range in which the gradient is non-zero (\cref{fig:hybnetapp:surrogate}~{\it bottom}) and the accuracy of the approximation (\cref{fig:hybnetapp:surrogate}~{\it top}).
This approach has two shortcomings: First, it has to be taken care that the gradient does not over- or underflow, which happens easily because of the exponential function.
Second, the gradient is only close to one for a very small range of inputs, leading to suboptimal convergence.
The authors of \cite{zenke_2018_1514} propose to use the fast sigmoid
\begin{equation}
\sigma_f(x,a) = (\frac{ax}{1+|ax|} + 1) / 2,
\end{equation}
a numerically more stable and faster to calculate approximation of the sigmoid (red and purple lines in \cref{fig:hybnetapp:surrogate}).
The gradient of the fast sigmoid,
\begin{equation}
\frac{\partial\sigma_f(x,a)}{\partial x} = \frac{1}{(|ax| + 1)^2}
\end{equation}
is in the range $[0, 1]$ with the maximum at $0$ for arbitrary steepness.
A similarly numerically stable, but less accurate approximation can be obtained by using the first or second order Taylor expansion of the sigmoid (yellow line in \cref{fig:hybnetapp:surrogate})
\begin{equation}
T(x, a) = (1/2 + ax/4 - (ax)^3/48).
\end{equation}
It is also possible to use a linear function
\begin{equation}
L(x, a) = ax/2 + 0.5
\end{equation}
 or $\relu$ to approximate the sigmoid (brown and pink lines in \cref{fig:hybnetapp:surrogate}).
All of the above have to be constraint to a range (typically $[0,\thresh]$ or $[-\thresh, \thresh]$, leading to constant gradients in a fixed range.
The functions themselves are only crude approximations of the original activation function.
The DIET-SNN surrogate \cite{rathi2020dietsnn} takes a mixed approach and defines only the gradient as a triangular function with the maximum at $0$, leading to linear gradients symmetrically around the threshold.

\begin{figure}[t]
\begin{center}
   \includegraphics[width=0.7\linewidth]{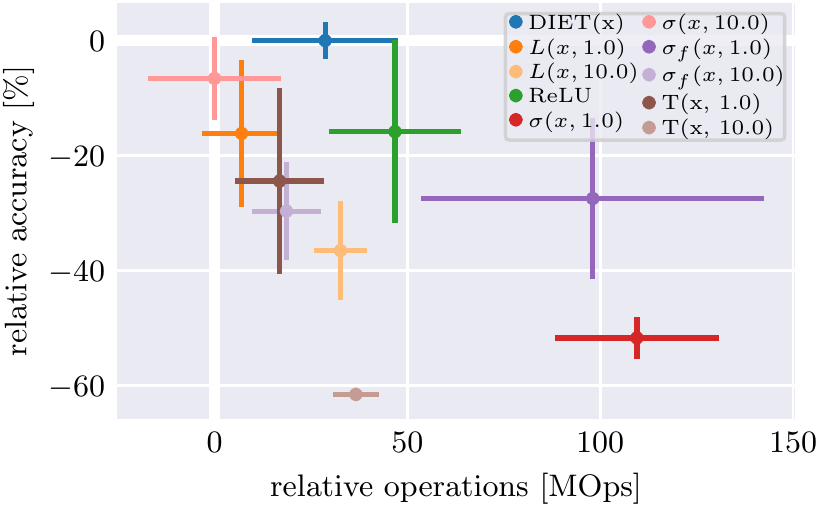}
\end{center}
   \caption{
   	Comparison of accuracy vs. operations for different surrogate gradients from Fig.~\ref{fig:hybnetapp:surrogate}.
Results are recorded with a hybrid DenseNet ($\growthfactor=16$, $\nouts=2$, $\nlayersperblock=2$, $\nblocks=2$)
on a \num{1800} samples, 4 class subset of the \nmnist{} dataset.
We report validation accuracy and number of operations, each relative to the best achieved result, \ie the best architecture is in the top-left.
The validation accuracy was chosen as to prevent optimizing hyperparameters on the test set.
}
\label{fig:hybnetapp:SurrogateComp}
\end{figure}

Comparing experimental results for  different surrogate gradient methods from the literature is difficult, because every paper
uses different network architectures.
On the other hand, treating the surrogate function as a hyperparameter to sweep over in conjunction with the
architecture hyperparameters like growth factor or training hyperparameters like the learning rate would lead to an intractable search space.
Therefore, we chose to evaluate the surrogate functions first, by choosing an average-sized architecture (hybrid DenseNet with
$\growthfactor=16$, $\nouts=2$, $\nlayersperblock=2$, $\nblocks=2$) and train on a \num{1800} samples, 4 class subset
of the \nmnist{} training dataset (see \cref{fig:hybnetapp:SurrogateComp}).
The best result in terms of accuracy is obtained with the 'diet' gradient from \cite{rathi2020dietsnn}, using a scale factor of 1.
A sigmoid with $a=10$ leads to a lower number of operations, but we decided that this does not compensate for the obtained loss in accuracy.

\subsection{Training with surrogate gradients}
\label{sec:surrogateGrad}

Our goal is to train the hybrid network end-to-end with backpropagation.
We use a surrogate gradient \cite{neftci_2019_36} for \cref{eq:spikes}, to overcome the problem that the analytical gradient is 0 almost everywhere in SNN layers.
Recently, multiple surrogate gradients have been proposed for SNN training.
We provide an experiment with different surrogate gradients on a small subsample of the \nmnist{} dataset in the appendix, with the result that the DIET surrogate gradient proposed in \cite{rathi2020dietsnn} gives the best results.
It is defined as
\begin{align}
\dfrac{\mathrm{d}\Theta(x)}{\mathrm{d}x} \approx
\begin{cases}
   \gradscale(\thresh - |x|),& \text{if } -\thresh \leq x \leq \thresh\\
    0,              & \text{otherwise}
\end{cases}
\end{align}
with a scaling factor $\gradscale$.

Similar to \cite{rathi2020dietsnn}, we train the threshold voltage and the membrane leakage with backpropagation.
Additionally, in the same way, we train the synaptic leakage.
For the leakages and the threshold, we use one global parameter per layer,
because \cite{rathi2020dietsnn} indicated that using one parameter for each neuron in each layer does not lead to better results.

\section{Detailed description of \nmnist{} networks}

Our experiments compared VGG, DenseNet and \DenseSep{} backbones, using different growth factors $\growthfactor \in \{8, 16\}$,
layers per block $\nlayersperblock \in \{2, 3\}$ and number of output intervals $\nouts \in \{1, 8\}$.
All neuron parameters (threshold and both leakages) are trained jointly with the weights.
They are initialized as $\thresh=1.$, $\taumem=\SI{20}{ms}$ and $\tausyn=\SI{1e-8}{ms}$.
The networks are rolled out for $\ntimesteps=\SI{150}{ms}$, with outputs at $\{24, 42, 60, 78, 96, 114, 132, 150\}~\si{ms}$ for $\nouts=8$.
The ANN head always consists of two stacks, where the first stack consists of one block and the second stack has three blocks.
A linear classification layer of shape (80, 10) is attached to the flattened output of the last layer in the last stack.

\section{Synaptic delays}

Synaptic delays are present in biological systems \cite{gonzales_2000_monkey} as well as on dedicated hardware \cite{nilsson_2020_delay},
and therefore have to be taken into account during training.
We implement delays by evaluating the maximum delay of out-going connections for each layer and initialize a ring buffer of this size to store the spiking activations.
With this approach, we save memory compared to simpler approaches, like storing the activations at the input of each layer or storing all activations over all time steps.
This allows training deeper networks, as the total available memory on GPU is the limiting factor for the network size.
As a side effect, this introduces an additional way to integrate information in time, see \cref{sec:timescales} and \cite{fischer2018}.

\section{Timescales}
\label{sec:timescales}
Our architecture has three intrinsic timescales: On the lowest level, the membrane potential integrates information over multiple time steps before emitting a spike.
This timescale is determined by the factors $\memleak$ and $\thresh$ in \cref{eq:neuronstate} and \cref{eq:spikes}.
As a rule of thumb, we observe that the weights in combination with the threshold are often such that only 1-3 spikes are needed to excite a neuron.
On the second level, there are the delayed connections in the SNN, transporting information forward in time.
As each connection has a delay of at least 1, the delay from input to output without skip connection is equal to the depth of the SNN.
Finally, the SNN can have multiple output intervals $\tout$, which enables the ANN to integrate information over the timescale of multiples of the depth of the network,
if the output intervals are separated by enough time steps.

\section{Optimizing SSD prior boxes}
\label{sec:priorBoxes}

In the SSD architecture \cite{Liu_2016}, each set of features is processed by
two convolutional layers that predict the bounding box coordinates relative to pre-defined prior boxes and the class for each prior box.
The first convolutional layer maps each spatial location to $n$ prior boxes and predicts a class score for each class and box.
The second convolutional layer has $4n$ output channels
and regresses the $x_0$, $y_0$, $x_1$ and $y_1$ coordinates to match each prior box as close as possible to the closest ground truth box.
Only prior boxes with an intersection-over-union (IoU) larger than a predefined threshold with a bounding box are matched with that box,
which naturally leads to a lot of bounding boxes being assigned to the background class.
Therefore, the loss is only calculated at a specific negative-positive ratio of all prior boxes, typically set to values aroung $3$.
Because the number of positive training samples is highly influenced by the quality of the prior boxes,
tuning the hyperparameters that define the prior boxes is of particular significance.

\begin{figure}[t]
\begin{center}
   \includegraphics[width=0.7\linewidth]{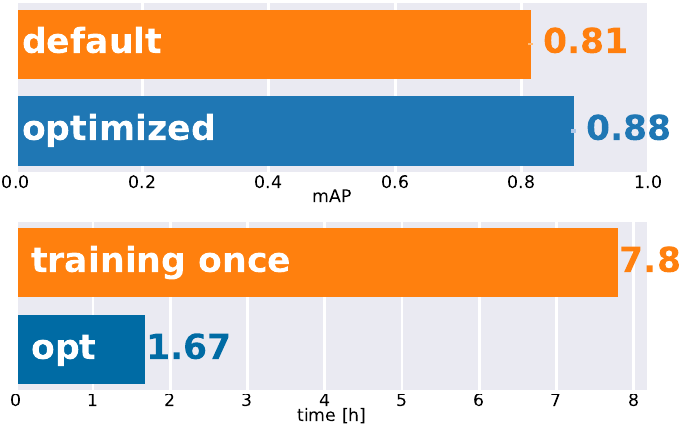}
\end{center}
   \caption{
{\it Top:} Mean average precision on \shapesEasy{} for the default prior boxes and the prior boxes found after our optimization.
{\it Bottom:} Time to train the network once and time to do the whole optimization with 2000 samples.
Our method finds better prior boxes at a fraction of the time of one training.
Usually, multiple training steps are needed to optimize the prior box hyperparameters.
}
\label{fig:hybnetapp:priorBoxOpt}
\end{figure}

\begin{figure}[t]
\begin{center}
   \includegraphics[width=0.35\linewidth]{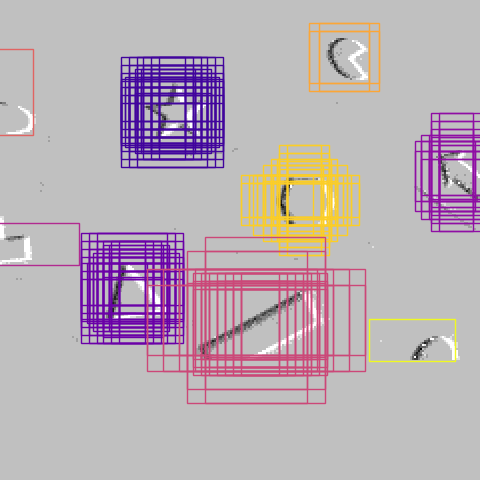}
   \includegraphics[width=0.35\linewidth]{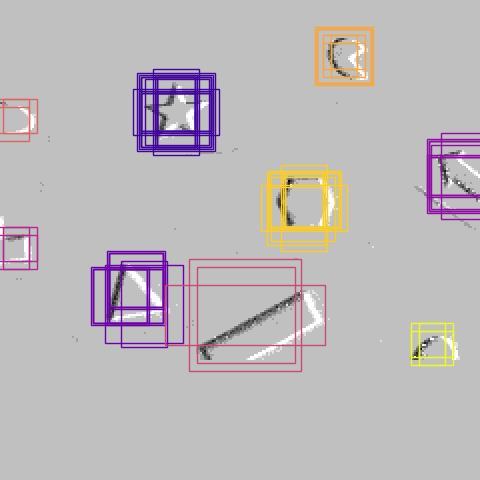}
\end{center}
   \caption{
\textit{Left:} Prior boxes \emph{before} optimization.
\textit{Right:} Prior boxes \emph{after} our optimization.
Each image shows all prior boxes that are matched to a ground truth target box during training.
After optimization, the prior boxes for smaller targets fit a lot better, particularly at the edge of the frame.
At the same time, there are less prior boxes for bigger targets,
which should be beneficial for post-processing methods like non-maximum suppression,
because less predictions have to be filtered.
}
\label{fig:hybnetapp:priorsVis}
\end{figure}

In the following, we are presenting a novel way to optimize the prior box hyperparameters of the SSD architecture without training the neural network.
The SSD architecture as described in \cref{sec:od} has multiple Detector Heads, whereas each head is defined by its prior boxes.
For each head $i$, the set of priors is described by the minimum size $\minsize$, maximum size $\maxsize$ and aspect ratio $\aspratio$
such that four prior boxes are defined with shapes:
$(\minsize,\minsize)$, $(\sqrt{\minsize\maxsize},\sqrt{\minsize\maxsize})$,
$(\minsize\cdot\aspratio,\frac{\minsize}{\aspratio})$, $(\frac{\minsize}{\aspratio},\minsize\cdot\aspratio)$.
We argue, that good prior boxes are defined by their overlap with bounding boxes during training.
As the ground truth training bounding boxes are available, we therefore do not have to optimize a loss defined by the network predictions,
but can define and optimize a function that describes the relationship between prior boxes and ground truth boxes.
This function should inlcude two objectives: First, it is important to have a high intersection over union (IoU) between ground truth boxes and prior boxes.
This can in practice be measured by the mean IoU over all combinations between training and ground truth bounding boxes.
However, this can lead to all prior boxes fitting well to only a subset of the training boxes.
Therefore we instead take the mean over the maximum IoU between a prior box and all ground truth boxes.
Second, to enforce that each prior box is matched to at least one ground truth box,
we also count the number of prior boxes with a maximum IoU over the detection threshold $\detectionthresh$.
We combine both objectives with a $2:1$ weight as
\begin{align}
\lossprior = (\meaniou + 2\npriorsthresh / \npriors) / 3
\end{align}
We chose to do the simplest approach to optimize this function and just sampled 2000 random points.
Generally, more intricate approaches like Bayesian Optimization would be possible as well, but the result from random search was already sufficient for our purposes.
To sample suitable points, we also consider the receptive field of each feature map:
Each prior box should be smaller than their respective receptive field
to ensure we only predict bounding boxes based on information that is available to the predictor.
We incorporate this by calculating the theoretical receptive field \cite{wenjie_2016} for each feature map
and setting the bounds for the maximum sizes to \SI{90}{\%} of the receptive field.
Additionally, we want to enforce that the minimum sizes grow monotonically with the receptive field and always be smaller than the maximum sizes.
We achieve this by using the maximum size of head $i-1$ as minimum size of head $i$.
For the results in all following sections, we optimized the hyperparameters by selecting four feature maps from our network with sizes $\{60,30,15,7\}$ and receptive fields $\{34,38,62,110\}$.
We sample 2000 random points and compare the prior boxes with the smallest loss to the default prior boxes of the SSD architecture in \cref{fig:hybnetapp:priorBoxOpt}.
The backbone is a DenseNet SNN architecture with the same parameters as the best model in \cref{sec:nmnist}.
Our mean average precision increases by 0.07 (averaged over two runs) compared to the default prior boxes.
Additionally, the whole optimization process takes about \SI{1.67}{h} for 2000 random samples, while training the network once for 50 epochs already takes \SI{7.8}{h}.
Our best result, ordered as $\minsize$, $\maxsize$, $\aspratio$ is $\{10,27,32,49\}$, $\{27,32,49,64\}$, $\{2.0,1.07,1.84,2.69\})$.
In \cref{fig:hybnetapp:priorsVis}, we compare the prior boxes used for training before and after optimization.
We find that smaller boxes, particularly at the edge, are matched much better.
There are less prior boxes for bigger ground truth boxes.
However, there are still enough well-matching boxes and therefore should not affect training performance.
Having less matches can even help to speed up inference, because less boxes have to be compared during non-maximum suppression.

Pseudo-code for the function to be optimized (minimized) is shown in \cref{lst:priorBoxes}.
This function can be straight-forwardly optimized using \eg Bayesian optimization.
\begin{lstlisting}[language=python,label=lst:priorBoxes,caption=Pseudocode to optimize prior boxes]
def priors_loss(hyperparameters):
    # feature maps are defined by backbone
    feature_map_sizes = get_feature_map_sizes(backbone)
    min_sizes, max_sizes, aspect_ratios = hyperparameters
    # generally non-differentiable
    prior_boxes = define_priors(min_sizes, max_sizes, aspect_ratios, feature_map_sizes)
    # normalized between 0 and 1
    ious_mean = calculate_mean_over_max_ious_per_prior(prior_boxes, ground_truth_boxes)
    n_priors_bigger_thresh = calculate_priors_bigger_than_thresh(prior_boxes, ground_truth_boxes)
    return -(ious_mean + 2*n_priors_bigger_thresh / len(prior_boxes)) / 3
\end{lstlisting}

\section{The \shapes{} dataset}

\begin{figure}[t]
\begin{center}
   \includegraphics[width=0.99\linewidth]{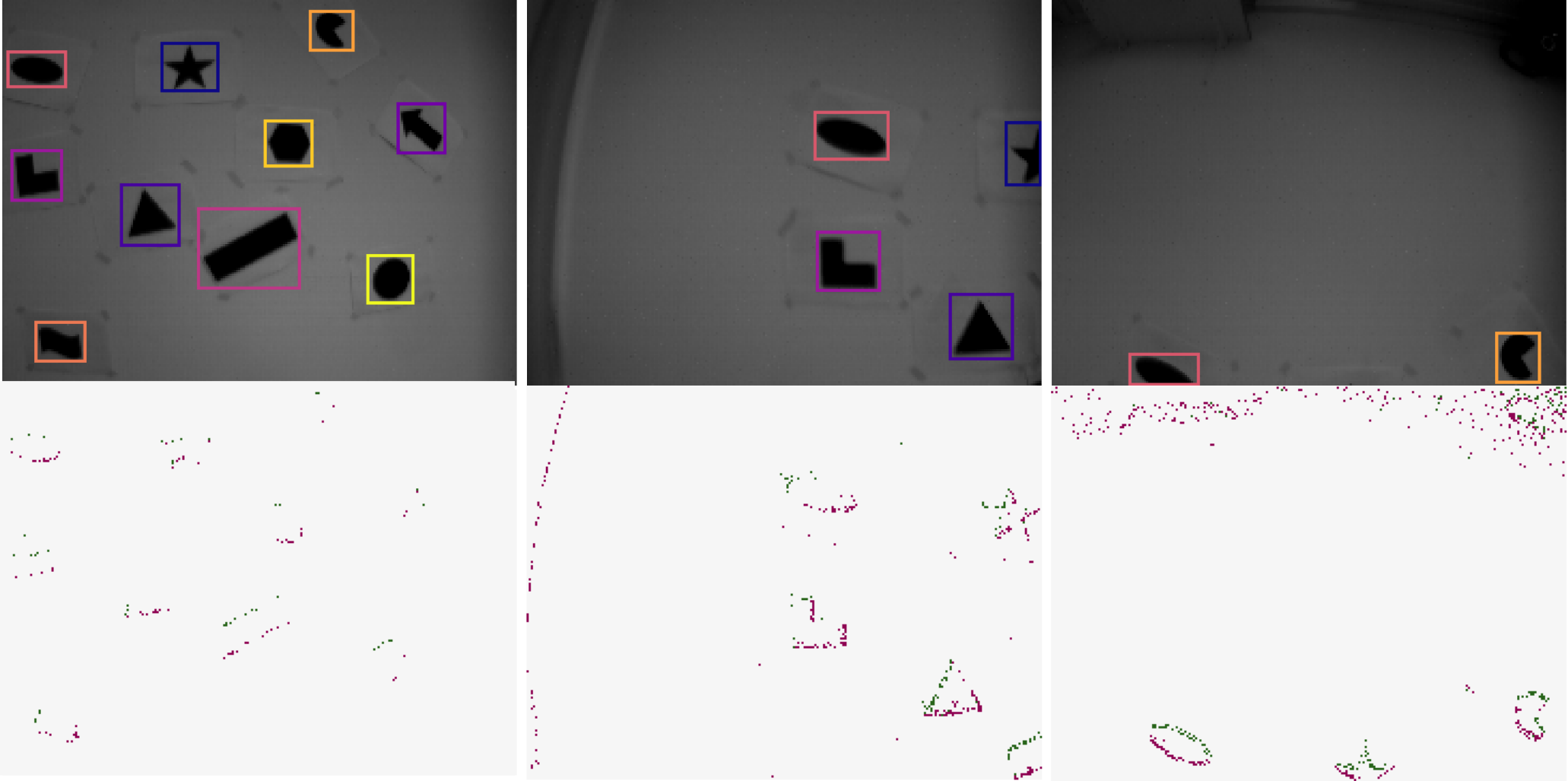}
\end{center}
   \caption{
Three samples from the  dataset with our ground truth bounding boxes.
{\it Top:} Grayscale frames used for labeling.
{\it Bottom:} Events accumulated over \SI{1}{ms}.
}
\label{fig:hybnetapp:shapesExample}
\end{figure}

The scene was recorded by moving a DAVIS240C \cite{brandli_2014_49} around
with varying speeds at a resolution of $180\times240$ pixels for a duration of $\SI{59.75}{s}$.
Three samples of the dataset are shown in \cref{fig:hybnetapp:shapesExample}.
While the shapes cannot overlap and there are no occlusions in the dataset,
it is still non-trivial to predict the bounding boxes from the accumulated event data,
as their appearance changes due to the camera movement.
For example, the triangle is barely visible on the left , while for the middle sample, it can be clearly identified.
Particularly difficult are those samples where shapes move in or out of the picture,
as in the sample in the middle or right.
The paper on which the shapes are printed on (middle),
as well as the background (middle and right) add additional noise to the data.
The task is to predict bounding boxes in a predefined time interval from the asynchronous event data.
As the whole dataset is small, we decided to provide two train/test splits together with the bounding boxes:
An easier split, called \shapesEasy{} where \SI{90}{\%} of the data is randomly assigned to the train set
and the other part is used as test set,
and a more difficult split \shapesHard{},
where only \SI{30}{\%} of the data is used in the train set.
We also considered splitting the sets in two time intervals,
such that the separation between train and test set is higher,
but due to the small size and the fact that only one scene is recorded,
we decided this split does not provide an additional benefit.

\subsection{Object detection performance without finetuning}
\label{sec:freeze}

\begin{figure}[t]
\begin{center}
   \includegraphics[width=0.7\linewidth]{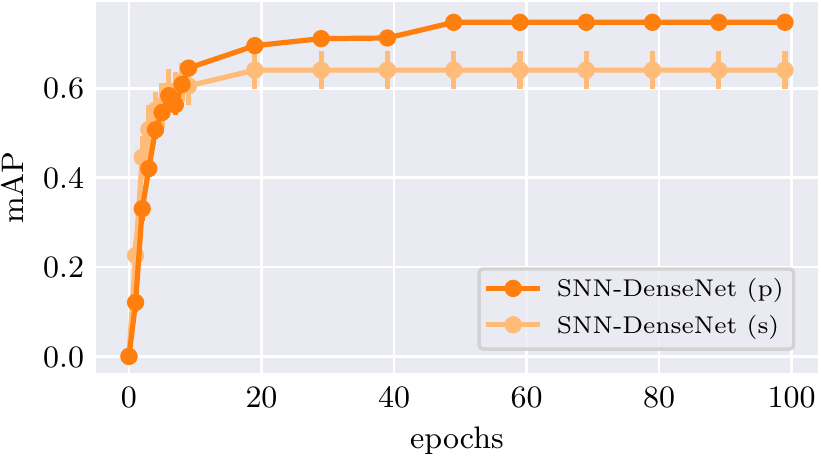}
\end{center}
   \caption{
Mean average precision over epochs for a hybrid network with frozen SNN backbone on \shapesHard{}.
The network with a pretrained SNN backbone improves over random features without further training of the backbone.
Mean and error of the mean over two trials.
}
\label{fig:hybnetapp:freeze}
\end{figure}

To investigate if the features learned for \nmnist{} are general enough to be helpful on other datasets,
we train only the ANN head of a network with two different weight initializations for the backbone.
One is initialized with random weights, while the other was pretrained with a different ANN head on \nmnist{}.
It can be seen in \cref{fig:hybnetapp:freeze} that the pretrained backbone reaches a significantly higher mAP than the random network.
The backbone learned features on the \nmnist{} dataset that are also useful for the \shapesEasy{} task.

\section{\shapes{} Details}
For all backbones, we take the architecture parameters from the best network in the \nmnist{} task.
For VGG and DenseNet, this is a network with $\growthfactor=16$, $\nlayersperblock=2$, $\nblocks=2$,
while the \DenseSep{} architecture has $\growthfactor=8$, $\nlayersperblock=3$, $\nblocks=2$.
For the ANN head, we choose a network consisting of two stacks, where the first stack has 1 block and the second stack has 3 blocks.
SSD heads are attached after each block of the last stack,
and additionally after the first convolution of the last stack, to enable a denser set of bounding boxes in space with a small spatial resolution.

\begin{figure}[t]
\begin{center}
   \includegraphics[width=0.7\linewidth]{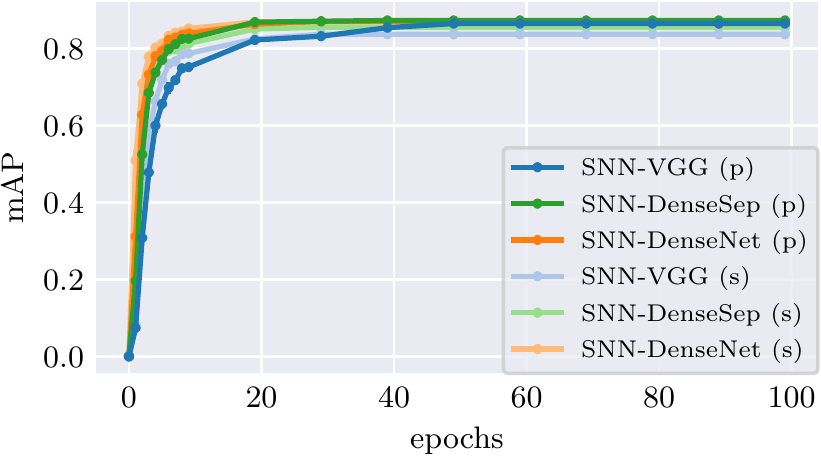}
\end{center}
   \caption{
Mean average precision (mAP) over the number of epochs for \shapesEasy{}.
Networks trained from scratch are marked with (s), pretrained networks with (p).
}
\label{fig:hybnetapp:shapesEasy}
\end{figure}

\begin{figure}[t]
\begin{center}
   \includegraphics[width=0.7\linewidth]{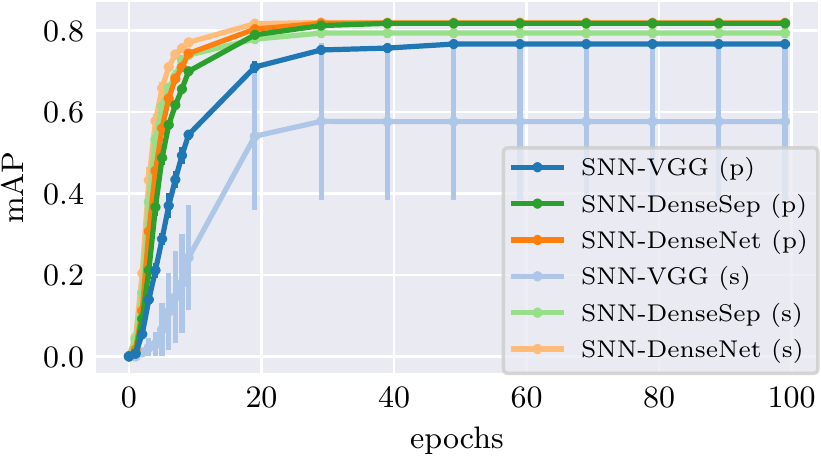}
\end{center}
   \caption{
Mean average precision (mAP) over the number of epochs for \shapesHard{}.
Networks trained from scratch are marked with (s), pretrained networks with (p).
}
\label{fig:hybnetapp:shapesHard}
\end{figure}

See \cref{tab:shapeseasy} and \cref{fig:hybnetapp:shapesEasy} for results on \shapesEasy{}.
See \cref{tab:shapeshard} and \cref{fig:hybnetapp:shapesHard} for results on \shapesHard{}.

\begin{table*}
\begin{center}
\begin{tabular}{lllll}
\toprule
architecture                                       & mAP                 & \#ops $[\mathrm{MOps}]$ & \#params & bw [\si{MB/s}] \\
\midrule
SNN DenseNet (s) & $86.98\pm0.69$ & $3000\pm130$ & 744939 & 3.52 \\
SNN DenseSep (s) & $85.57\pm0.58$ & $1393\pm1.3$ & 409553 & 2.48\\
SNN VGG (s) & $83.78\pm0.44$ & $1361\pm61$ & 336037 & 3.20\\
ANN DenseNet & $63.40\pm0.55$ & $14690$ & 151040 & 864\\
ANN DenseSep & $64.54\pm0.18$ & 7841 & 238344 & 1210\\
ANN VGG & $62.85\pm0.44$ & 83064 & 336016 & 1380\\
SNN DenseNet (p) & $87.09\pm0.20$ & $2679.0\pm7.1$ & 744939 & 11.5\\
SNN DenseSep (p) & $87.37\pm0.51$ & $1398.9\pm2.3$ & 409553 & 11.0\\
SNN VGG (p) & $86.70\pm0.36$ & $1611\pm32$ & 336037 & 7.49\\
\bottomrule
\end{tabular}
\caption{
Results on \shapesEasy{} (4 trials each).
We report the mean and error of the mean.
Networks trained from scratch are marked with (s), pretrained networks with (p).
\#ops are the number of operations, \#params the number of parameters of the model and bw is the bandwidth between SNN backbone and ANN head.
}
\label{tab:shapeseasy}
\end{center}

\begin{center}
\begin{tabular}{lllll}
\toprule
architecture                            & mAP           & \#ops $[\mathrm{MOps}]$ & \#params & bw [\si{MB/s}] \\
\midrule
SNN DenseNet (s) & $82.0\pm1.0$  & $2790\pm50$ & 744939 & 2.68\\
SNN DenseSep (s) & $79.4\pm1.2$  & $1384.6\pm1.0$ & 409553 & 2.22\\
SNN VGG (s) & $58\pm19$        & $1230\pm89$   & 336037 & 1.82\\
ANN DenseNet & $54.20\pm0.64$ & $14690$ & 151040 & 864\\
ANN DenseSep & $56.78\pm0.24$ & 7841 & 238344 & 1210\\
ANN VGG & $57.72\pm0.36$                & 83064 & 336016 & 1380\\
SNN DenseNet (p) & $81.76\pm0.53$ & $2696.5\pm3.3$ & 744939 & 16.1\\
SNN DenseSep (p) & $81.74\pm0.61$ & $1392.3\pm1.0$ & 409553 & 10.7\\
SNN VGG (p) & $76.66\pm0.58$ & $1303\pm21$ & 336037 & 14.8\\
\bottomrule
\end{tabular}
\caption{
Results on \shapesHard{} (4 trials each).
We report the mean and error of the mean.
Networks trained from scratch are marked with (s), pretrained networks with (p).
\#ops are the number of operations, \#params the number of parameters of the model and bw is the bandwidth between SNN backbone and ANN head.
}
\label{tab:shapeshard}
\end{center}
\end{table*}

\section{Regularization for homeostasis}

During training we face the problem that sometimes the network does not start to train, because no spikes reach the last layer.
As the surrogate gradient of $\Theta(\state - \thresh)$ is only non-zero around the threshold, neurons that are not spiking, also cannot learn and therefore, cannot learn to spike.
We could track this back to the weight initialisation, that sometimes leads to a suboptimal inital state.
We tried to tackle this problem by including homeostasis during training, \ie introducing a loss that increases activity
\begin{equation}
\lossspikes = 0.5(\sum_{tln}S_{tln} - \targetrate)^2,
\end{equation}
with indices time step $t$, layer $l$ and neuron $n$.
Experimentally, we did not find an improvement.
This can be explained by looking at the calculated weight update for a single-layer network.
\begin{align}
\Delta w_{ij} = -\lr(\sum_{t}S_{tj} - \targetrate)(\left.\frac{\partial \Theta}{\partial x}\right|_{x=\state_i-\thresh}\sum^t_{k=0}\memleak^kI_{j-k}),
\end{align}
\ie the update is also always 0 and we cannot force the network to an average target rate if the activation is too low in the beginning.

\section{Membrane potential initialization}
We tried to utilize the initial membrane potential to increase training accuracy, because by default,
the initial membrane potentials $\state_0$ are a free parameter during training.
In some works \cite{kugele_2020_effproc,rueckauer2017} it is set to 0 for each sample, which covers the case where the initial membrane potentials is reset after each prediction.
To be independent from the starting conditions, it is possible to start with random values for all $\state_0$.
This can also be seen as data augmentation.
We tried to draw initial membrane potentials from a normal distribution
\begin{equation}
\state_0 \sim \mathcal{N}(\thresh / 2,\,\thresh / 10)
\end{equation}
for each sample during training and also experimented with testing with random initialization vs.~initialization from 0.
We found in practice that often the results on the validation set are worse in both cases and therefore decided to stick with initialization to 0 for training and during validation and test time.

\section{Bandwidth details}
We introduce an additional $\mathrm{L}_1$ loss on the activations of the output layer of the SNN,
\begin{equation}
L_\mathrm{out} = \frac{\lambdaout}{BTWH} \sum_{b,i,x,y} |S_{\mathrm{out},i,b,x,y}|
\end{equation}
with a scaling factor $\lambdaout$, the batch size $B$, total simulation time $T$, and width $W$ and height $H$ of the output layer.
It is important to notice that we do not need to take the absolute value,
because all activations are binary, \ie positive by definition.
For the same reason, the loss is normalized to 1.

\section{Parameters for classification tasks}
\cref{tab:densenetarch,tab:denseseparch,tab:vggarch,tab:ffarch} list all used architectures for the classification tasks.
SNN-ANNs have more parameters than ANN-ANNs because we also count the neuron parameters $\thresh$, $\memleak$, $\synleak$ that are shared across layers.

\begin{table*}
\begin{center}
\begin{tabular}{lrrrlr}
\toprule
backbone &  $\growthfactor$ & $\nlayersperblock$ & $\nouts$ &    SNN & $n_{params}$ \\
\midrule
     densenet &   8 &    2 &     1 &  False &    64854 \\
     densenet &   8 &    2 &     1 &   True &    64875 \\
     densenet &   8 &    2 &     8 &  False &    64854 \\
     densenet &   8 &    2 &     8 &   True &    64875 \\
     densenet &   8 &    3 &     1 &  False &   127014 \\
     densenet &   8 &    3 &     1 &   True &   127041 \\
     densenet &   8 &    3 &     8 &  False &   127014 \\
     densenet &   8 &    3 &     8 &   True &   127041 \\
     densenet &  16 &    2 &     1 &  False &   256414 \\
     densenet &  16 &    2 &     1 &   True &   256435 \\
     densenet &  16 &    2 &     8 &  False &   256414 \\
     densenet &  16 &    2 &     8 &   True &   256435 \\
     densenet &  16 &    3 &     1 &  False &   503998 \\
     densenet &  16 &    3 &     1 &   True &   504025 \\
     densenet &  16 &    3 &     8 &  False &   503998 \\
     densenet &  16 &    3 &     8 &   True &   504025 \\
\bottomrule
\end{tabular}
\caption{
All DenseNet architectures for the classification tasks.
}
\label{tab:densenetarch}
\end{center}
\end{table*}

\begin{table*}
\begin{center}
\begin{tabular}{llllll}
\toprule
backbone &  $\growthfactor$ & $\nlayersperblock$ & $\nouts$ &    SNN & $n_{params}$ \\
\midrule
 densenet\_sep &   8 &    2 &     1 &  False &    60454 \\
 densenet\_sep &   8 &    2 &     1 &   True &    60487 \\
 densenet\_sep &   8 &    2 &     8 &  False &    60454 \\
 densenet\_sep &   8 &    2 &     8 &   True &    60487 \\
 densenet\_sep &   8 &    3 &     1 &  False &   117774 \\
 densenet\_sep &   8 &    3 &     1 &   True &   117819 \\
 densenet\_sep &   8 &    3 &     8 &  False &   117774 \\
 densenet\_sep &   8 &    3 &     8 &   True &   117819 \\
 densenet\_sep &  16 &    2 &     1 &  False &   237374 \\
 densenet\_sep &  16 &    2 &     1 &   True &   237407 \\
 densenet\_sep &  16 &    2 &     8 &  False &   237374 \\
 densenet\_sep &  16 &    2 &     8 &   True &   237407 \\
 densenet\_sep &  16 &    3 &     1 &  False &   464014 \\
 densenet\_sep &  16 &    3 &     1 &   True &   464059 \\
 densenet\_sep &  16 &    3 &     8 &  False &   464014 \\
 densenet\_sep &  16 &    3 &     8 &   True &   464059 \\
\bottomrule
\end{tabular}
\caption{
All \DenseSep{} architectures for the classification tasks.
}
\label{tab:denseseparch}
\end{center}
\end{table*}

\begin{table*}
\begin{center}
\begin{tabular}{llllll}
\toprule
backbone &  $\growthfactor$ & $\nlayersperblock$ & $\nouts$ &    SNN & $n_{params}$ \\
\midrule
          vgg &   8 &    2 &     1 &  False &    50174 \\
          vgg &   8 &    2 &     1 &   True &    50195 \\
          vgg &   8 &    2 &     8 &  False &    50174 \\
          vgg &   8 &    2 &     8 &   True &    50195 \\
          vgg &   8 &    3 &     1 &  False &   125582 \\
          vgg &   8 &    3 &     1 &   True &   125609 \\
          vgg &   8 &    3 &     8 &  False &   125582 \\
          vgg &   8 &    3 &     8 &   True &   125609 \\
          vgg &  16 &    2 &     1 &  False &   198254 \\
          vgg &  16 &    2 &     1 &   True &   198275 \\
          vgg &  16 &    2 &     8 &  False &   198254 \\
          vgg &  16 &    2 &     8 &   True &   198275 \\
          vgg &  16 &    3 &     1 &  False &   498830 \\
          vgg &  16 &    3 &     1 &   True &   498857 \\
          vgg &  16 &    3 &     8 &  False &   498830 \\
          vgg &  16 &    3 &     8 &   True &   498857 \\
\bottomrule
\end{tabular}
\caption{
All VGG architectures for the classification tasks.
}
\label{tab:vggarch}
\end{center}
\end{table*}

\begin{table*}
\begin{center}
\begin{tabular}{llllll}
\toprule
backbone &  $\growthfactor$ & $\nlayersperblock$ & $\nouts$ &    SNN & $n_{params}$ \\
\midrule
        ffann &   8 &    2 &     1 &  False &   711394 \\
        ffsnn &   8 &    2 &     1 &   True &   691389 \\
\bottomrule
\end{tabular}
\caption{
All baseline architectures for the classification tasks.
}
\label{tab:ffarch}
\end{center}
\end{table*}

\end{document}